%% file: acl_latex.tex
\pdfoutput=1

\documentclass[11pt]{article}

\usepackage[]{acl}

\usepackage{times}
\usepackage{latexsym}
\usepackage[utf8]{inputenc}
\usepackage[T1]{fontenc}
\usepackage{CJKutf8}
\usepackage[english]{babel}
\usepackage{makecell}
\usepackage{graphicx}
\usepackage{hyperref}
\usepackage{multirow} 
\usepackage{float} 
\usepackage{amsmath} 
\usepackage{adjustbox} 
\usepackage{verbatim} 
\usepackage{kotex} 
\usepackage{bbm}
\usepackage{booktabs}
\usepackage{arydshln} 
\usepackage{textcomp} 
\usepackage{dblfloatfix} 
\usepackage[T1]{fontenc}

\usepackage[utf8]{inputenc}

\usepackage{microtype}

\title{Don't be a Fool: Pooling Strategies in Offensive Language Detection\\ from User-Intended Adversarial Attacks}

\author{Seunguk Yu, Juhwan Choi \and Youngbin Kim \\
  Chung-Ang University, Republic of Korea, Seoul \\
  \texttt{seungukyu@gmail.com, \{gold5230, ybkim85\}@cau.ac.kr} \\
}

\begin{document}
\maketitle
\begin{abstract}
\textit{Warning: this paper contains expressions that may offend the readers.} \\
\\
Offensive language detection is an important task for filtering out abusive expressions and improving online user experiences. However, malicious users often attempt to avoid filtering systems through the involvement of textual noises. In this paper, we propose these evasions as user-intended adversarial attacks that insert special symbols or leverage the distinctive features of the Korean language. Furthermore, we introduce simple yet effective pooling strategies in a layer-wise manner to defend against the proposed attacks, focusing on the preceding layers not just the last layer to capture both offensiveness and token embeddings. We demonstrate that these pooling strategies are more robust to performance degradation even when the attack rate is increased, \textit{without} directly training of such patterns. Notably, we found that models pre-trained on clean texts could achieve a comparable performance in detecting attacked offensive language, to models pre-trained on noisy texts by employing these pooling strategies.
\end{abstract}

\section{Introduction}

\input{01_introduction}

\section{Related Work}

\input{02_relatedwork}

\section{Method}

\input{03_method}

\section{Experiment}

\input{Table_3}

\input{04_experiments}

\section{Discussion}

\input{05_discussion}

\section{Conclusion}

\input{06_conclusion}

\section*{Limitations}
Despite our efforts to define diverse types of adversarial attacks, there is room for undefined attacks from real-world situations, which may have an unexpected impact on the model performance. Additionally, although we used some language-independent attacks such as inserting special symbols, most of the proposed attacks were based on the characteristics of the Korean language. Therefore, it is necessary to determine whether layer-wise pooling strategies can effectively handle attacked offensive expressions written in other languages for cross-lingual applications. The introduced pooling strategies offer significant flexibility for models in other languages, requiring simple modifications to the model structure.

\section*{Ethics Statement}
Given our use of offensive representations to describe the proposed attacks, we have included a disclaimer at the beginning of this paper. From the existing offensive language datasets, potential biases regarding race, gender, political issues, and other factors might have been inherent in our experiments. This should be considered when developing our research or expanding it to other languages.

\section*{Acknowledgements}
This research was supported by Basic Science Research Program through the National Research Foundation of Korea (NRF) funded by the Ministry of Education (NRF-2022R1C1C1008534), and Institute for Information \& communications Technology Planning \& Evaluation (IITP) through the Korea government (MSIT) under Grant No. 2021-0-01341 (Artificial Intelligence Graduate School Program, Chung-Ang University).

\bibliography{anthology,custom}
\bibliographystyle{acl_natbib}

\vspace{1\baselineskip}

\appendix

\input{_A-Appendix}

\input{_B-Appendix}

\input{_C-Appendix}

\end{document}

%% file: 01_introduction.tex
As the internet becomes an important part of our lives, the prevalence of offensive language on online platforms, particularly social media, has become a serious concern \cite{zampieri-etal-2019-predicting}. Deep learning models for filtering offensive languages have been proposed to address this problem. However, malicious users have consistently found ways to avoid them. One such way is the deliberate insertion of additional typographical errors or substitution of certain characters with visually similar alternatives \cite{kurita2019towards, wu-etal-2018-decipherment}.

Despite numerous studies on this phenomenon in English, there has been a comparatively limited exploration in Korean, which is a low-resource language characterized by distinct linguistic features \cite{sahoo-etal-2023-prejudice, kim2021commonsense}. As the Korean communities also suffer from the use of abusive language and cyberbullying \cite{mccurry22south, saengprang2021cyberbullying, yi2020cyber, jun2020study}, it is desirable to investigate the evasion tactics utilized by malicious users and to formulate them. While recent studies have discussed how to avoid offensive language detection \cite{cho2021google, kim2021trkic, sang2021morpho}, their definitions are ambiguous, and no clear solutions have been proposed to defend against the evasions.

In this paper, we propose the evasion methods as user-intended adversarial attacks and incorporate them into offensive language from the perspective of malicious users. Our proposed attacks are grounded in prevalent forms that can be found in offensive language online, and reflect the distinct features of Korean language, wherein a single character can be further subdivided \cite{song2006Korean}. We test the proposed attacks on existing models for offensive language detection, and the results reveal that the performance declines as the rate of the proposed attacks increases.

Furthermore, we introduce simple yet effective pooling strategies in a layer-wise manner to defend against the proposed attacks. Motivated by the exploration of the impact of each layer in a pre-trained language model \cite{oh-etal-2022-dont, jawahar-etal-2019-bert}, we selectively integrate useful features for the attacked offensive language across all layers\footnote{In our paper, we denote offensive language as `attacked' when an evasion method is applied to it, viewed from the perspective of malicious users.}. The attacked texts have some changes in the tokens used, differing from the original texts. Therefore, we implement pooling strategies to ensure that the model captures not only high-level features but also low-level features, which are related to offensiveness and token embeddings, respectively. This simple modification enriches the understanding of the attacked offensive language, enhancing the robustness of the model against user-intended adversarial attacks \textit{without} directly training of such patterns.

The contributions of our study are as follows:

\begin{itemize}
\item We propose user-intended adversarial attacks that are often associated with offensive language online from the perspective of malicious users. These attacks are performed by inserting special symbols or leveraging the distinctive features of Korean.

\item We introduce pooling strategies in a layer-wise manner to selectively utilize all layers rather just than the last layer. This approach achieves a notable performance when employed to a model pre-trained on clean texts, \textit{without} directly training of the attacks.

\item We demonstrate the effectiveness of layer-wise pooling strategies by assigning distinct weights to each layer and employing them to the model depending on the nature of the pre-trained texts. We especially note the efficacy of first-last pooling and max pooling when the attacks are involved in offensive language.
\end{itemize}

%% file: 02_relatedwork.tex
\subsection{Adversarial Attacks}
Adversarial attacks involve perturbed input data that confuses the trained model, whereas a situation in which the model consistently makes predictions regardless of the nature of input data is referred to as defending \cite{goyal2023survey}. Previous studies have explored this based on word-level substitutions \cite{jin2020bert, ren2019generating}, and others also have explored them on character-level. We propose adversarial attacks that utilize not only character-level but also smaller-scale alternations tailored to the features of Korean language. Such attacks are commonly observed in the context of offensive languages in various online communities \cite{cho2021google, sang2021morpho}.

TextBugger \cite{li2019textbugger} was an early study that focused on character-level alternations, such as replacing characters with visually similar ones (e.g. replacing the alphabet `o' with the number `0'). Other studies have suggested simple leetspeaks that utilize symbols that resemble the alphabet \cite{aggarwal2022analyzing}, or adversarial attacks that are not easily detected visually, such as transforming Latin characters into similar-looking Cyrillic characters \cite{wolff2020attacking}. Although several studies also have explored visually undetectable attacks \cite{bajaj2023homochar, boucher2022bad, kim2021hypocrite}, we consider a more realistic attack scenario that can occur online from the perspective of malicious users.

\subsection{Korean Offensive Language Datasets}
Owing to the increasing demand for online content in Korean language and the growing threat of cyberbullying, previous studies have introduced offensive language datasets collecting comments from diverse resources such as online news, communities, and YouTube.

BEEP! \cite{moon-etal-2020-beep} was a pioneering study that utilizes hate speech prevalent in news comments. KoLD \cite{jeong-etal-2022-kold} and K-MHaS \cite{lee-etal-2022-k} specified the target group of the offensive language. Subsequently, KODOLI \cite{park-etal-2023-feel} provided labels that refine the degree of offensiveness, and built upon these efforts, K-HATERS \cite{park2023k} was built to incorporate the strengths of the preceding datasets.

Although numerous datasets have been proposed, there is still a lack of definition for adversarial attacks that are frequently involved in offensive language, and how to defend against them. In this study, we focus on introducing pooling strategies for defending against these attacks, \textit{without} directly training attacked offensive langauge.

%% file: 03_method.tex
\subsection{User-Intended Adversarial Attacks}

We present adversarial attacks designed to target offensive languages from the perspective of malicious users. By referring to existing offensive language datasets, we define frequently occurring attack types. These attacks are categorized into three groups: \textsc{Insert}, \textsc{Copy}, and \textsc{Decompose}. Examples of each attack type are listed in Table \ref{table_1}.

First, \textsc{Insert} involves adding incomplete Korean character forms, which are often used online without significant meaning. For example, `ㅋㅋ' (equivalent `lol' or `lmao' in English) is a commonly used and somewhat meaningless string frequently used in online communications. In this case, \textsc{Insert}\_\textit{zz} is performed by inserting the string at a specific location within the word, as in real situations. Other types of \textsc{Insert} also include unnecessary spaces or special symbols.

\input{Table_1}

\input{Table_2}

The following two types of user-intended adversarial attacks take advantage of the distinct features of the Korean language; a single character must have an initial sound, a middle sound, and an optional final sound \cite{song2006Korean}. For example, in the expression `쓰레기 같은' in Table \ref{table_1}, the character `쓰' has only the initial and middle sounds, whereas the character `같' has all three sounds.

Second, \textsc{Copy} utilizes the distinctive features, copying one of the three sounds from the selected character to the other character. For example, the character `레' from the expression `쓰레기 같은' has the initial and middle sounds of `ㄹ' and `ㅔ'. In this case, \textsc{Copy}\_\textit{initial} is performed by copying the initial sound of that character `ㄹ' to the final sound of the preceding character `쓰'. Consequently, `쓰' is transformed into `쓸', leading to the attacked expression `쓸레기 같은'.

Finally, \textsc{Decompose} also utilizes the unique characteristics, isolating the final sound of the selected character or breaking down the character itself. For example, the single character `쓰' from the expression `쓰레기 같은' has the initial and middle sounds of `ㅆ' and `ㅡ'. In this case, \textsc{Decompose}\_\textit{all} is performed by breaking down the character, resulting in the sounds being independent of that character. Consequently, `쓰' is transformed into `ㅆㅡ', leading to the attacked expression `ㅆㅡ레기 같은'. Further details and examples of all user-intended adversarial attacks are provided in Appendix \ref{appendix_a}.

\subsection{Layer-Wise Pooling Strategies}

In standard text classification tasks, pre-trained models such as BERT are fined-tuned to the target domain. This is based on the assumption that the $\left[ CLS \right]$ token from the last layer effectively captures the sentence representation \cite{devlin2019bert}. However, we notice inconsistencies in the predictions of existing models regarding the proposed attacks\footnote{The results for this experiment are included in Table \ref{table_3}.}. Consequently, we conclude that this information alone is insufficient for detecting the attacked offensive language.

When using perturbed text to a trained model, the tokenization results differ from those of the original text, as shown in Table \ref{table_2}. By involving special symbols or exploiting the distinctive features of Korean, we observe that even if the text had the same meaning to human readers, the tokenized outputs differ significantly\footnote{The model used for tokenization in here is $\text{BERT}_{\textit{clean}}$.}. Therefore, we do not rely on the information only from the last layer but utilize the preceding layers, which focus more on token embeddings \cite{ma2019universal}. This also reflects the previous finding that meaningful information for a certain task can be captured in the preceding layers \cite{oh-etal-2022-dont}.

We extend pooling strategies in a layer-wise manner, allowing us the flexibility to utilize text representations from all layers. Denoting the $\left[ CLS \right]$ token of the $N$th layer as $h_{N}^{\textit{cls}}$, we introduce four pooling strategies that optionally consider the $\left[ CLS \right]$ tokens from all the layers $h_{1}^{\textit{cls}}, ..., h_{N}^{\textit{cls}}$.

\textbf{Mean, Max Pooling}: We apply mean pooling utilizing the $L^{1}$ norm, which averages all $\left[ CLS \right]$ tokens from all the layers, and max pooling utilizing the $L^{\infty}$ norm, which takes a max-over-time operation on the values corresponding to each dimension from all $\left[ CLS \right]$ tokens.

When the dimension of $\left[ CLS \right]$ token is $M$, and all the values of $m$th dimension of $\left[ CLS \right]$ tokens from all the layers are concatenated and denoted by $h_{\textit{all}}^{m}$, these two poolings are defined as follows:
\begin{align}\label{eq12}
pool_{\textit{mean}} &= mean(h_{1}^{\textit{cls}}, ..., h_{N}^{\textit{cls}}), \\
pool_{\textit{max}} &= max(h_{\textit{all}}^{1}), ...,  max(h_{\textit{all}}^{M}),
\end{align}

\textbf{Weighted Pooling}: We apply weighted pooling utilizing a learnable parameter that determines the importance of each layer. Through adaptive incorporation of the layers, we train weights that selectively capture both offensiveness and token embeddings, initializing all weights to zero.

When the $w_{i}$ represents the weight of each layer and $\alpha_{i}$ represents its softmax distribution, the weighted pooling is defined as follows:
\begin{align}\label{eq3}
pool_{\textit{weighted}} &= \sum_{i=1}^{N}{\alpha_{i}h_{i}^{\textit{cls}}},
\end{align}

\textbf{First-last Pooling}: We apply first-last pooling utilizing $\left[ CLS \right]$ token from the first layer. Rather than considering all the layers, we focus on leveraging information from layers directly associated with offensiveness and token embeddings.
\begin{align}\label{eq4}
pool_{\textit{first\text{-}last}} &= h_{N}^{\textit{cls}} + h_{1}^{\textit{cls}},
\end{align}

\begin{figure}[t!]
    \centerline{\includegraphics[width=\columnwidth]{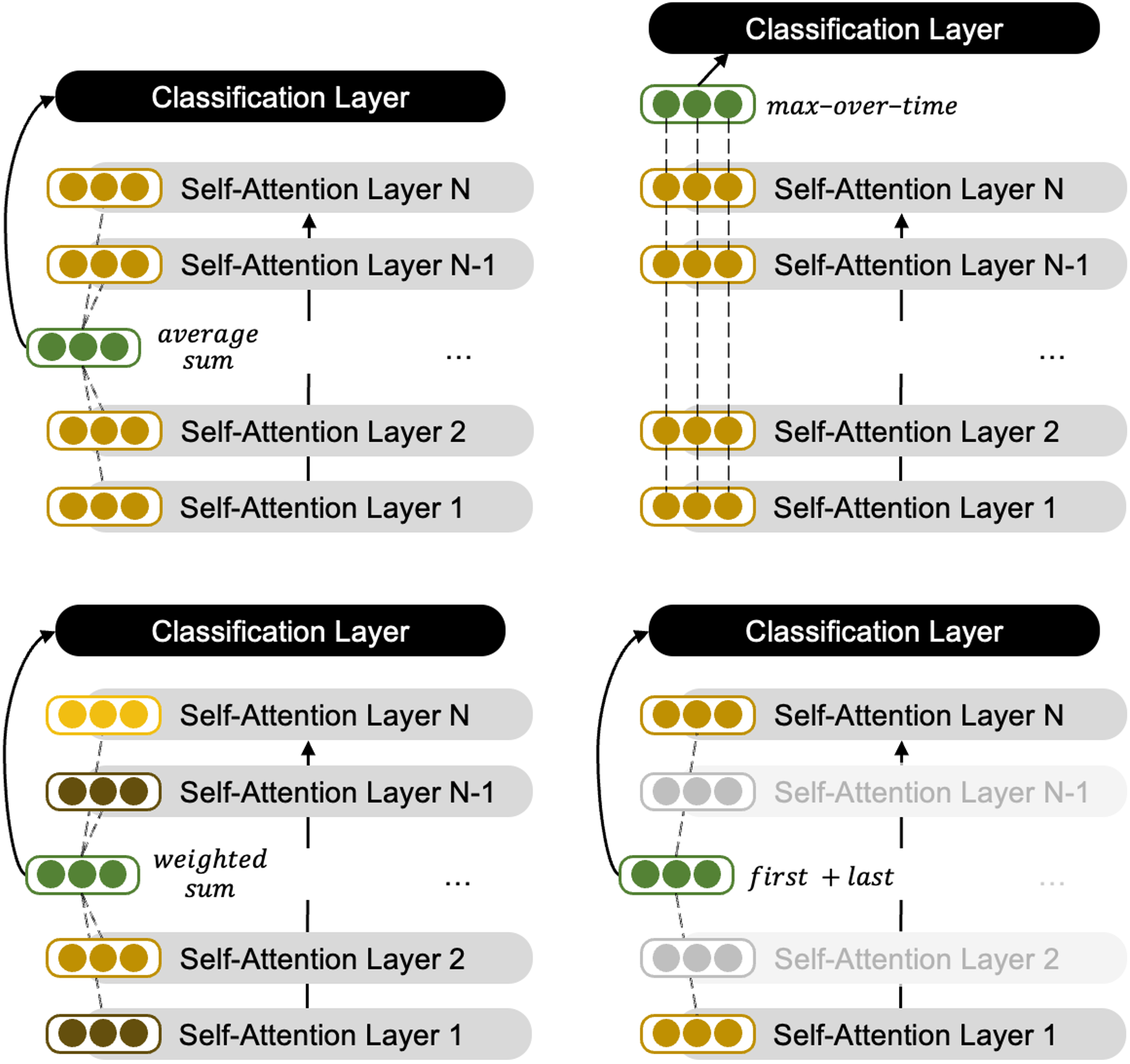}}
    \caption{Layer-wise pooling strategies that selectively use $\left[ CLS \right]$ tokens from all layers. From the upper left, there are mean, max, weighted, and first-last pooling.}
    \label{figure_1} 
\end{figure}

The layer-wise pooling strategies described above are illustrated in Figure \ref{figure_1}. We conducted experiments to verify the robustness of these strategies for detecting offensive languages that reflect user-intended adversarial attacks.

%% file: Table_1.tex
\begin{table}[t!]
\centering
\small
\begin{tabular}{l|l}
\begin{tabular}[c]{@{}l@{}}User-Intended\\ Adversarial Attacks\end{tabular} & Text Examples                    \\ \hline
\underline{original text}                                                         & \underline{쓰레기 같은} (piece of shit) \\
\textsc{Insert}\_\textit{zz}                                                                  & 씈ㅋㅋㅋ레기 같은                   \\
\textsc{Insert}\_\textit{space}                                                               & 쓰 레기 같은                     \\
\textsc{Insert}\_\textit{special}                                                             & 쓰@레기 같은                     \\
\textsc{Copy}\_\textit{initial}                                                               & 쓸레기 같은                      \\
\textsc{Copy}\_\textit{middle}                                                              & 쓰레에기 같은                     \\
\textsc{Copy}\_\textit{final}                                                              & 쓰레기 가튼                      \\
\textsc{Decompose}\_\textit{final}                                                         & 쓰레기 가ㅌ은                     \\
\textsc{Decompose}\_\textit{all}                                                            & ㅆㅡ레기 같은                    
\end{tabular}
\caption{Text examples of user-intended adversarial attacks with three categories: \textsc{Insert}, \textsc{Copy}, and \textsc{Decompose}. There are various attacks that involve special symbols or exploit the distinctive features of Korean.}
\label{table_1} 
\end{table}

%% file: Table_2.tex
\begin{table}[t!]
\centering
\small
\begin{tabular}{l|l}
\begin{tabular}[c]{@{}l@{}}User-Intended\\ Adversarial Attacks\end{tabular} & Tokenized Examples                    \\ \hline
\underline{original text}                                                         & \underline{쓰레기, 같, \#\#은} (piece of shit) \\
\textsc{Insert}\_\textit{zz}                                                                  & [UNK], 같, \#\#은                   \\
\textsc{Insert}\_\textit{space}                                                               & 쓰, 레, \#\#기, 같, \#\#은                     \\
\textsc{Insert}\_\textit{special}                                                             & 쓰, @, 레, \#\#기, 같, \#\#은                     \\
\textsc{Copy}\_\textit{initial}                                                               & 쓸, \#\#레기, 같, \#\#은                      \\
\textsc{Copy}\_\textit{middle}                                                              & 쓰, \#\#레, \#\#에, \#\#기, 같, \#\#은                     \\
\textsc{Copy}\_\textit{final}                                                              & 쓰레기, 가, \#\#튼                      \\
\textsc{Decompose}\_\textit{final}                                                         & 쓰레기, 가, \#\#ㅌ, \#\#은                     \\
\textsc{Decompose}\_\textit{all}                                                            & [UNK], 같, \#\#은                    
\end{tabular}
\caption{Tokenized examples of user-intended adversarial attacks. Although the texts have the same meaning, the tokens are represented differently.}
\label{table_2} 
\end{table}

%% file: Table_3.tex
\begin{table*}[t!]
\begin{adjustbox}{max width=\textwidth}
\begin{tabular}{l|ccc|cccl|cccl|cccl}
\hline
\specialrule{1pt}{0pt}{0pt}
\multicolumn{1}{c|}{\multirow{2}{*}{Model}}                                                      & \multicolumn{3}{c|}{Original}                                                      & \multicolumn{4}{c|}{30\% Attacked}                                                                                     & \multicolumn{4}{c|}{60\% Attacked}                                                                                     & \multicolumn{4}{c}{90\% Attacked}                                                                                     \\ \cline{2-16} 
\multicolumn{1}{c|}{}                                                                            & P                         & R                         & F1                         & P                         & R                         & \multicolumn{1}{c|}{F1}    & \multicolumn{1}{c|}{$\Delta atk$} & P                         & R                         & \multicolumn{1}{c|}{F1}    & \multicolumn{1}{c|}{$\Delta atk$} & P                         & R                         & \multicolumn{1}{c|}{F1}    & \multicolumn{1}{c}{$\Delta atk$} \\ \hline
\specialrule{1pt}{0pt}{0pt}
BiLSTM                                                                                           & 71.83                     & 68.80                     & 69.81                      & 70.84                     & 66.23                     & \multicolumn{1}{c|}{67.38} & -3.48\%                           & 68.97                     & 62.82                     & \multicolumn{1}{c|}{63.67} & \textbf{-8.79\%}                           & 69.32                     & 61.64                     & \multicolumn{1}{c|}{62.25} & \textbf{-10.82\%}                         \\ \hline
BiGRU                                                                                              & \multicolumn{1}{l}{71.32} & \multicolumn{1}{l}{65.71} & \multicolumn{1}{l|}{66.91} & \multicolumn{1}{l}{70.40} & \multicolumn{1}{l}{63.32} & \multicolumn{1}{l|}{64.26} & -3.96\%                           & \multicolumn{1}{l}{68.84} & \multicolumn{1}{l}{60.31} & \multicolumn{1}{l|}{60.52} & -9.55\%                           & \multicolumn{1}{l}{68.05} & \multicolumn{1}{l}{58.83} & \multicolumn{1}{l|}{58.48} & -12.59\%                         \\ \hline
$\text{BERT}_{\textit{clean}}$                                                                                  & 79.81                     & 77.79                     & 78.64                      & 79.51                     & 73.35                     & \multicolumn{1}{c|}{75.19} & -4.38\%                           & 77.74                     & 66.38                     & \multicolumn{1}{c|}{67.96} & -13.58\%                          & 76.14                     & 62.01                     & \multicolumn{1}{c|}{62.44} & -20.60\%                         \\ \hline
$\text{BERT}_{\textit{multi}}$                                                                                  & 76.57                     & 70.75                     & 72.38                      & 76.44                     & 66.33                     & \multicolumn{1}{c|}{67.87} & -6.23\%                           & 76.30                     & 60.90                     & \multicolumn{1}{c|}{60.87} & -15.90\%                          & 76.07                     & 57.61                     & \multicolumn{1}{c|}{55.78} & -22.93\%                         \\ \hline
\specialrule{1pt}{0pt}{0pt}
$\text{BERT}_{\textit{clean}} + mean$                                                                           & 78.57                     & 79.06                     & 79.01                      & 79.41                     & 73.97                     & \multicolumn{1}{c|}{75.70} & -4.18\%                           & 77.15                     & 66.97                     & \multicolumn{1}{c|}{68.62} & -13.15\%                          & 74.90                     & 62.45                     & \multicolumn{1}{c|}{63.09} & -20.14\%                         \\ \hline
$\text{BERT}_{\textit{clean}} + max$                                                                            & 78.51                     & 78.81                     & 78.65                      & 78.47                     & 74.03                     & \multicolumn{1}{c|}{75.54} & -3.95\%                           & 77.80                     & 66.66                     & \multicolumn{1}{c|}{68.29} & -13.17\%                          & 76.79                     & 61.51                     & \multicolumn{1}{c|}{61.72} & -21.52\%                         \\ \hline
$\text{BERT}_{\textit{clean}} + weighted$                                                                       & 79.93                     & 78.50                     & 79.14                      & 79.19                     & 73.70                     & \multicolumn{1}{c|}{75.43} & -4.68\%                           & 77.14                     & 67.62                     & \multicolumn{1}{c|}{69.33} & -12.39\%                          & 75.44                     & 63.66                     & \multicolumn{1}{c|}{64.85} & -18.05\%                         \\ \hline
$\text{BERT}_{\textit{clean}} + first\text{-}last$                                                                          & 79.05                     & 79.37                     & \textbf{79.21}                      & 78.89                     & 75.85                     & \multicolumn{1}{c|}{\textbf{77.02}} & \textbf{-2.76\%}                           & 77.58                     & 69.38                     & \multicolumn{1}{c|}{\textbf{71.21}} & \textbf{-10.09\%}                          & 76.08                     & 64.33                     & \multicolumn{1}{c|}{\textbf{65.49}} & \textbf{-17.32\%}                         \\ \hline
\specialrule{1pt}{0pt}{0pt}
$\text{BERT}_{\textit{noise}}$                                                                                  & 80.64                     & 78.88                     & 79.64                      & 80.67                     & 75.42                     & \multicolumn{1}{c|}{77.17} & -3.10\%                           & 78.44                     & 69.42                     & \multicolumn{1}{c|}{\textbf{71.33}} & -10.43\%                          & 76.46                     & 65.55                     & \multicolumn{1}{c|}{\textbf{66.96}} & -15.92\%                         \\ \hline
\begin{tabular}[c]{@{}l@{}}$\text{Ensemble}_{\textit{hard}}$\\ ($\text{BERT}_{\textit{clean}}$ + $\text{BERT}_{\textit{noise}}$)\end{tabular} & \multicolumn{1}{l}{81.63} & \multicolumn{1}{l}{79.42} & \multicolumn{1}{l|}{80.36} & \multicolumn{1}{l}{81.57} & \multicolumn{1}{l}{75.03} & \multicolumn{1}{l|}{77.04} & -4.13\%                           & \multicolumn{1}{l}{80.47} & \multicolumn{1}{l}{68.60} & \multicolumn{1}{l|}{70.60} & -12.14\%                          & \multicolumn{1}{l}{78.68} & \multicolumn{1}{l}{64.24} & \multicolumn{1}{l|}{65.35} & -18.67\%                         \\ \hline
\begin{tabular}[c]{@{}l@{}}$\text{Ensemble}_{\textit{soft}}$\\ ($\text{BERT}_{\textit{clean}}$ + $\text{BERT}_{\textit{noise}}$)\end{tabular} & 81.52                     & 79.53                     & \textbf{80.38}                      & 81.54                     & 75.29                     & \multicolumn{1}{c|}{\textbf{77.25}} & -3.89\%                           & 80.27                     & 68.86                     & \multicolumn{1}{c|}{70.87} & -11.83\%                          & 78.47                     & 64.42                     & \multicolumn{1}{c|}{65.58} & -18.41\%                         \\ \hline
DeBERTaV3                                                                                  & 82.55                     & 78.70                     & 80.17                      & 81.85                     & 74.33                     & \multicolumn{1}{c|}{76.48} & -4.60\%                           & 80.14                     & 68.47                     & \multicolumn{1}{c|}{70.44} & -12.13\%                          & 78.41                     & 63.96                     & \multicolumn{1}{c|}{64.99} & -18.93\%                         \\ \hline
\specialrule{1pt}{0pt}{0pt}
\end{tabular}
\end{adjustbox}
\caption{Experimental results of offensive language detection when a certain ratio of user-intended adversarial attacks are involved. P, R, and F1 represent macro precision, recall, and f1-score, respectively. $\Delta atk$ represents the performance drop in the f1-score as the attacks are involved in the same model.}
\label{table_3} 
\end{table*}

%% file: 04_experiments.tex
\subsection{Datasets}
We collected both the KoLD \cite{jeong-etal-2022-kold} and K-HATERS \cite{park2023k} datasets and divided them into train, validation, and test sets by stratifying their labels. We randomly shuffled and split them in the ratio of 8:1:1. We set the attack rates to 30\%, 60\%, and 90\%, corrupting a portion of the words in a sentence. The details of how the attacks were carried out are in Appendix \ref{appendix_a}.

\subsection{Baselines}
We validate the effectiveness of the layer-wise pooling strategies with the baselines, which are presented below. The experimental details including hyperparameters and metrics are reported in Appendix \ref{appendix_b}.
\begin{itemize}

\item BiLSTM: This model addresses the long-term dependency problem by remembering only the information in need \cite{schuster1997bidirectional}. It was built by stacking two LSTMs, and the forward and backward $\left[ CLS \right]$ tokens from the last layer were combined and passed through the classification layer.

\item BiGRU: This model is derived from the BiLSTM and further evolved by reducing the training parameters through the selective utilization of gates \cite{cho2014learning}. Its configurations are the same as BiLSTM.

\item $\text{BERT}_{\textit{clean}}$: This model follows the BERT \cite{devlin2019bert} structure, which is built on a self-attention mechanism with masked language modeling. It was pre-trained on preprocessed texts in Korean \cite{park2021klue}. The $\left[ CLS \right]$ token from the last layer is passed through the classification layer.

\item $\text{BERT}_{\textit{multi}}$: This model is pre-trained on multilingual data, including Korean~\cite{conneau2020unsupervised}. The model structure is based on RoBERTa~\cite{liu2019roberta}, while the remaining configurations are the same as $\text{BERT}_{\textit{clean}}$.

\item $\text{BERT}_{\textit{noise}}$: This model is pre-trained on noisy texts in Korean, such as online comments~\cite{lee2020kcbert}. Its configurations are the same as $\text{BERT}_{\textit{clean}}$.

\item $\text{Ensemble}_{\textit{hard}}$, $\text{Ensemble}_{\textit{soft}}$: These models utilize both the $\text{BERT}_{\textit{clean}}$ and $\text{BERT}_{\textit{noise}}$, employing voting methods from ensemble techniques. Hard voting is conducted through a majority vote, but soft voting occurs in the cases of tied votes. Soft voting averages the prediction probabilities of each model.

\item DeBERTaV3: This model employs gradient-disentangled embeddings to enhance the efficiency of pre-trianing when incorporating replaced token detection along with DeBERTa~\cite{he2023debertav3, he2021deberta}. We used the model fine-tuned in Korean\footnote{https://huggingface.co/team-lucid/deberta-v3-base-korean}.
\end{itemize}

%% file: 05_discussion.tex
\subsection{Experimental Results}

The performances of the models when exposed to user-intended adversarial attacks are presented in Table \ref{table_3}. Each of the best performances from the layer-wise pooling strategies and the baselines in the original, and 30\%, 60\%, and 90\% attacked are highlighted in bold.

All baselines, including ensemble models and even a recent model like DeBERTaV3, were susceptible to the proposed attacks. We observed that the BERT-based models consistently outperformed RNN-based models in terms of the f1-score across all attack rates. Under original and 30\% attacked, employing ensemble models with soft voting yielded the best scores, achieving f1-scores of 80.38 and 77.25. As the attack rates increased to 60\% and 90\%, using a single model pre-trained on noisy texts proved to be the most effective, achieving f1 scores of 71.33 and 66.96, respectively.

However, ensemble models require twice computational resources for both training and inference stages compared to a single model. In the case of $\text{BERT}_{\textit{noise}}$, a large amount of noisy texts is required, raising concerns regarding its adaptability when inference with attacked input types is not encountered during the pre-training stage.

\input{Table_4}

When applying layer-wise pooling strategies to $\text{BERT}_{\textit{clean}}$, we found that the performances were improved in almost all attack rates. They only need to train an additional parameter equal to the size of all layers (e.g. 12 for BERT-based models), or no parameters are required. Furthermore, they are robust as the attack rate increases compared to models with no pooling strategies.

The average performances when exposed to proposed attacks across all attack rates are presented in Table \ref{table_4}. All layer-wise pooling strategies exhibited robustness against attacks compared to their absence, except for $\text{BERT}_{\textit{clean}} + max$, which exhibited a slight performance drop. Moreover, even $\text{BERT}_{\textit{clean}} + first\text{-}last$, which only incorporates information from the first layer \textit{without} any parameters or training noisy texts, showed comparable performance to $\text{BERT}_{\textit{noise}}$ across all attack rates that were pre-trained on noisy texts.

\subsection{Focus on Performance Drop}

\begin{figure}[b!]
    \centerline{\includegraphics[width=\columnwidth]{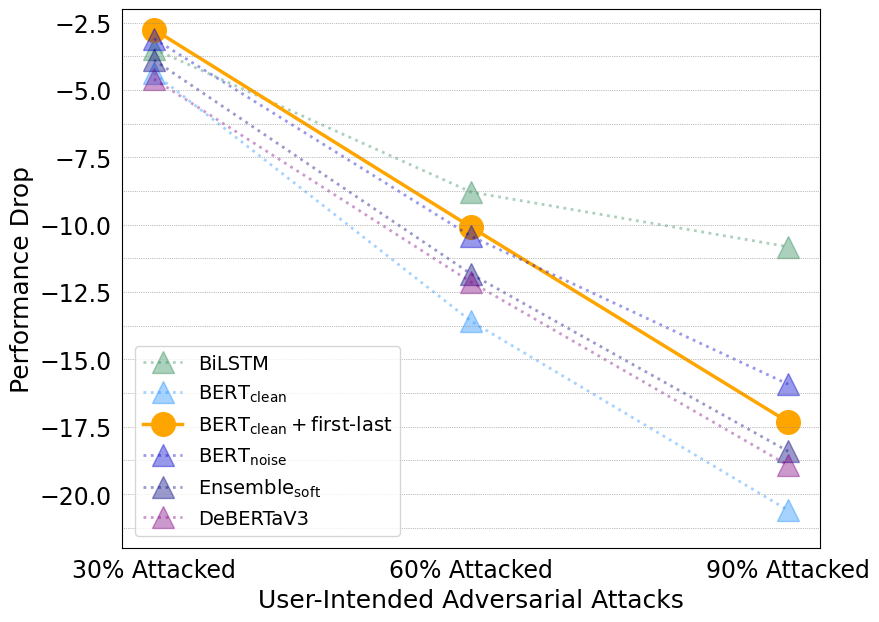}}
    \caption{Degree to which the f1-scores of the models decrease with the attack rates 30\%, 60\%, and 90\%. We selected several baseline models and $\text{BERT}_{\textit{clean}} + first\text{-}last$ for the comparison.}
    \label{figure_2} 
\end{figure}

The degree to which the f1-scores of the models decreased with the attack rates is shown in Figure \ref{figure_2}. We found that BiSLTM exhibited relatively modest performance degradation across the models, which is depicted by the light green triangles. Despite the modest decrease, the offensive language detection scores of RNN-based models were not as good as that of the BERT-based models because of the limitations of themselves.

Among the BERT-based models, $\text{BERT}_{\textit{clean}}$, which was pre-trained on clean texts and is depicted by the light blue triangles, exhibited the largest performance degradation. However, $\text{BERT}_{\textit{clean}} + first\text{-}last$, which applied a simple layer-wise pooling strategy to the model and is depicted by the orange circles, successfully mitigated performance degradation by 1.62\%, 3.49\%, and 3.28\% at each attack rate, respectively, achieving an average performance degradation mitigation of 2.79\%.

These results are similar or even better to those of $\text{BERT}_{\textit{noise}}$ or $\text{Ensemble}_{\textit{soft}}$, which used noisy texts in a pre-training stage and are depicted by the blue and deep blue triangles, respectively. Therefore, we found that the model pre-trained on clean texts with a simple pooling strategy can achieve a certain level of performance, or even be more robust compared to the model pre-trained on noisy texts in defending against user-intended adversarial attacks. It was even more robust than the recent model DeBERTaV3, which is depicted by the purple triangles.

\subsection{Focus on Layer Weights}

\begin{figure}[t!]  
    \centerline{\includegraphics[width=6cm]{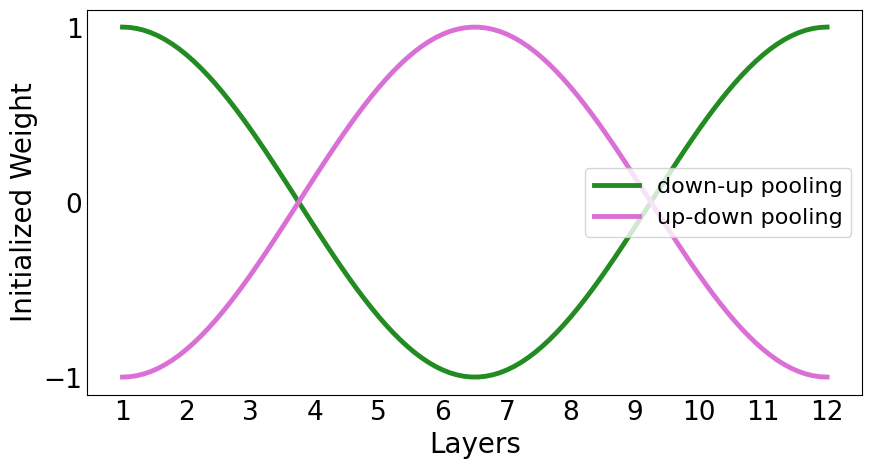}}
    \caption{Initialized weights for each of the down-up and up-down poolings. Each strategy shares the shape of a cosine function but varies in the range on the x-axis depending on the layers to be focused.}
    \label{figure_3} 
\end{figure}

\input{Table_5}

\input{Table_6}

Additionally, we conducted experiments to determine whether useful information could be captured not only in the first and last layer, but also in the layers close to these two layers of the model. We hypothesized that layers close to the last layer would capture the offensiveness that determine the text representations, whereas layers close to the first layer would capture the token embeddings that determine the degree to which a sentence contains textual attacks. We set all weights to zero for the weighted pooling, however, in this experiment, we assigned distinct weights to all layers.

The weights for the down-up and up-down poolings are shown in Figure \ref{figure_3}. We assigned relatively high weights to the layers close to the last and first layers, while assigning low weights to the middle layers. This strategy is referred to as down-up pooling, as its weight graph moves down then up. The green line at the cosine function within the range $[0, 2\pi]$ represents the initial weights for down-up pooling. In contrast, we assigned relatively low weights to the layers close to the last and first layers, while assigning high weights to the middle layers. This strategy is referred to as up-down pooling, as its weight graph moves up then down. The pink line at the cosine function within the range $[\pi, 3\pi]$ represents that of the up-down pooling.

The performances of the models for the distinct layer weights are presented in Table \ref{table_5}. Down-up pooling outperformed for all attack rates and exhibited more robustness against performance degradation compared to up-down pooling. This demonstrates that it is more important to weigh the layers close to the last and first layers than those of the middle for attacked offensive languages. In summary, the strategy that focuses on layers capable of capturing offensiveness and token embeddings yields a better performance.

Nevertheless, neither of these pooling strategies performed as well as the $\text{BERT}_{\textit{clean}}$ or the model that employed the first-last pooling. This indicates that while information close to the last and first layers can be beneficial, there are some layers among them that hinder the performance of the model, revealing that only the last and first layers are the most helpful layers when detecting offensive language with user-intended adversarial attacks.

\subsection{Beyond the Limits: Broader Application\\
\;\;\;\;\;\;\;\;of Layer-Wise Pooling Strategies}

We employed the layer-wise pooling strategies not only to $\text{BERT}_{\textit{clean}}$, a model pre-trained on clean texts, but also to models that utilize noisy texts or employ alternative methods in the pre-training stage. The performances of the layer-wise pooling strategies for both the $\text{BERT}_{\textit{noise}}$ and DeBERTaV3 are presented in Table \ref{table_6}. In $\text{BERT}_{\textit{noise}}$, first-last pooling only performed well on the performance degradation, but the strategy exhibiting more robustness to detection metrics was max pooling.

In the case of $\text{BERT}_{\textit{clean}}$, because it did not directly utilize noisy texts, first-last pooling performed well for attacked offensive language, which focuses most on offensiveness and token embeddings from the sentence. However, in the case of $\text{BERT}_{\textit{noise}}$, where noisy texts were pre-trained into the model weights before the application of pooling strategies, max pooling proved to be the most effective on detection metrics by selecting prominent features among the information to predict the labels invariant to small changes or disturbances.

The results for DeBERTaV3 showed that it performed better with only the model itself when there were no attacks or 30\% attacked, while performed better with pooling strategies when the attack rate was increased. We found that the distinct pre-training process of DeBERTaV3, different from the general BERT, was capable of handling a certain level of textual attacks. However, the introduced pooling strategy could prove beneficial in scenarios where the attack rate increases. 

The average performances using max pooling and first-last pooling, which had the most impact on each model are presented in Table \ref{table_7}. For the model $\text{BERT}_{\textit{clean}}$ and DeBERTaV3, first-last pooling improved the f1-score by 2.72 and 0.15 and prevented a performance degradation of 2.8\% and 0.38\%, respectively. In particular, the naive BERT model with the pooling strategies outperformed the ensemble models, and even surpassed the recent model DeBERTaV3 with the pooling strategies. It suggests that the introduced pooling strategies can be employed for a simple structure that solely relies on naive BERT, enabling efficient detection of attacked offensive language without the necessity for the training of noisy texts and strategic pre-training.

In the case of $\text{BERT}_{\textit{noise}}$, max pooling improved its f1-score by 2.26, reaching a relatively high score of 74.08. In terms of performance degradation, first-last pooling exhibited a performance degradation of 2.81\%. Thus, we confirmed that implementing the pooling strategies on a model pre-trained with noisy texts resulted in more robust and improved detection of attacked offensive language exceeded others, such as pooling strategies on a model pre-trained with clean texts. The performances of these pooling strategies are remarkable considering that they do not require additional parameters or direct training on noisy texts.

\input{Table_7}

%% file: Table_4.tex
\begin{table}[t!]
\centering
\small
\begin{adjustbox}{max width=\textwidth}
\begin{tabular}{l|cl}
\hline
\specialrule{1pt}{0pt}{0pt}
\multicolumn{1}{c|}{\multirow{2}{*}{Model}}                                                      & \multicolumn{2}{c}{Average}                                           \\ \cline{2-3} 
\multicolumn{1}{c|}{}                                                                            & \multicolumn{1}{c|}{F1}             & \multicolumn{1}{c}{$\Delta atk$} \\ \hline
\specialrule{1pt}{0pt}{0pt}
BiLSTM                                                                                           & \multicolumn{1}{c|}{64.43}          & \textbf{-7.69\%}                          \\ \hline
BiGRU                                                                                            & \multicolumn{1}{l|}{61.08}          & -8.70\%                          \\ \hline
$\text{BERT}_{\textit{clean}}$                                                                                  & \multicolumn{1}{c|}{68.52}          & -12.85\%                         \\ \hline
$\text{BERT}_{\textit{multi}}$                                                                                  & \multicolumn{1}{c|}{61.50}          & -15.02\%                         \\ \hline
\specialrule{1pt}{0pt}{0pt}
$\text{BERT}_{\textit{clean}} + mean$                                                                           & \multicolumn{1}{c|}{69.13}          & -12.49\%                         \\ \hline
$\text{BERT}_{\textit{clean}} + max$                                                                            & \multicolumn{1}{c|}{68.51}          & -12.88\%                         \\ \hline
$\text{BERT}_{\textit{clean}} + weighted$                                                                       & \multicolumn{1}{c|}{69.86}          & -11.70\%                         \\ \hline
$\text{BERT}_{\textit{clean}} + first\text{-}last$                                                                          & \multicolumn{1}{c|}{\textbf{71.24}} & \textbf{-10.05\%}                \\ \hline
\specialrule{1pt}{0pt}{0pt}
$\text{BERT}_{\textit{noise}}$                                                                                  & \multicolumn{1}{c|}{\textbf{71.82}} & -9.81\%                 \\ \hline
\begin{tabular}[c]{@{}l@{}}$\text{Ensemble}_{\textit{hard}}$\\ ($\text{BERT}_{\textit{clean}}$ + $\text{BERT}_{\textit{noise}}$)\end{tabular} & \multicolumn{1}{l|}{70.99}          & -11.64\%                         \\ \hline
\begin{tabular}[c]{@{}l@{}}$\text{Ensemble}_{\textit{soft}}$\\ ($\text{BERT}_{\textit{clean}}$ + $\text{BERT}_{\textit{noise}}$)\end{tabular} & \multicolumn{1}{c|}{71.23}          & -11.37\%                         \\ \hline
DeBERTaV3                                                                                  & \multicolumn{1}{c|}{70.63}          & -11.88\%                         \\ \hline
\specialrule{1pt}{0pt}{0pt}
\end{tabular}
\end{adjustbox}
\caption{Average from the experimental results of offensive language detection when a certain ratio of user-intended adversarial attacks are involved.}
\label{table_4} 
\end{table}

%% file: Table_5.tex
\begin{table*}[t!]
\begin{adjustbox}{max width=\textwidth}
\begin{tabular}{l|ccc|cccl|cccl|cccl}
\hline
\specialrule{1pt}{0pt}{0pt}
\multicolumn{1}{c|}{\multirow{2}{*}{Model}}                                                      & \multicolumn{3}{c|}{Original}                                                      & \multicolumn{4}{c|}{30\% Attacked}                                                                                     & \multicolumn{4}{c|}{60\% Attacked}                                                                                     & \multicolumn{4}{c}{90\% Attacked}                                                                                     \\ \cline{2-16} 
\multicolumn{1}{c|}{}                                                                            & P                         & R                         & F1                         & P                         & R                         & \multicolumn{1}{c|}{F1}    & \multicolumn{1}{c|}{$\Delta atk$} & P                         & R                         & \multicolumn{1}{c|}{F1}    & \multicolumn{1}{c|}{$\Delta atk$} & P                         & R                         & \multicolumn{1}{c|}{F1}    & \multicolumn{1}{c}{$\Delta atk$} \\ \hline
\specialrule{1pt}{0pt}{0pt}
$\text{BERT}_{clean}$                                                                                  & 79.81                     & 77.79                     & 78.64                      & 79.51                     & 73.35                     & \multicolumn{1}{c|}{75.19} & -4.38\%                           & 77.74                     & 66.38                     & \multicolumn{1}{c|}{67.96} & -13.58\%                          & 76.14                     & 62.01                     & \multicolumn{1}{c|}{62.44} & -20.60\%                         \\ \hline
\specialrule{1pt}{0pt}{0pt}
$\text{BERT}_{clean} + down\text{-}up$                                                                                              & \multicolumn{1}{l}{79.80} & \multicolumn{1}{l}{77.64} & \multicolumn{1}{l|}{\textbf{78.55}} & \multicolumn{1}{l}{80.02} & \multicolumn{1}{l}{73.05} & \multicolumn{1}{l|}{\textbf{75.02}} & \textbf{-4.49\%}                           & \multicolumn{1}{l}{77.20} & \multicolumn{1}{l}{65.70} & \multicolumn{1}{l|}{\textbf{67.15}} & \textbf{-14.51\%}                           & \multicolumn{1}{l}{76.61} & \multicolumn{1}{l}{61.84} & \multicolumn{1}{l|}{\textbf{62.19}} & \textbf{-20.82\%}                         \\ \hline
$\text{BERT}_{clean} + up\text{-}down$                                                                                              & \multicolumn{1}{l}{80.35} & \multicolumn{1}{l}{77.10} & \multicolumn{1}{l|}{78.36} & \multicolumn{1}{l}{79.95} & \multicolumn{1}{l}{71.67} & \multicolumn{1}{l|}{73.73} & -5.90\%                           & \multicolumn{1}{l}{78.59} & \multicolumn{1}{l}{65.28} & \multicolumn{1}{l|}{66.67} & -14.91\%                           & \multicolumn{1}{l}{76.81} & \multicolumn{1}{l}{60.87} & \multicolumn{1}{l|}{60.81} & -22.39\%                         \\ \hline
\specialrule{1pt}{0pt}{0pt}
$\text{BERT}_{clean} + first\text{-}last$                                                                          & 79.05                     & 79.37                     & \textbf{79.21}                      & 78.89                     & 75.85                     & \multicolumn{1}{c|}{\textbf{77.02}} & \textbf{-2.76\%}                           & 77.58                     & 69.38                     & \multicolumn{1}{c|}{\textbf{71.21}} & \textbf{-10.09\%}                          & 76.08                     & 64.33                     & \multicolumn{1}{c|}{\textbf{65.49}} & \textbf{-17.32\%}                         \\ \hline
\specialrule{1pt}{0pt}{0pt}
\end{tabular}
\end{adjustbox}
\caption{Experimental results of offensive language detection when using down-up and up-down pooling strategies. They are initialized along with the cosine function, assigning distinct weights to the layers depending on whether they focus more on offensiveness and token embeddings.}
\label{table_5} 
\end{table*}

%% file: Table_6.tex
\begin{table*}[t!]
\begin{adjustbox}{max width=\textwidth}
\begin{tabular}{l|ccc|cccl|cccl|cccl}
\hline
\specialrule{1pt}{0pt}{0pt}
\multicolumn{1}{c|}{\multirow{2}{*}{Model}}                                                      & \multicolumn{3}{c|}{Original}                                                      & \multicolumn{4}{c|}{30\% Attacked}                                                                                     & \multicolumn{4}{c|}{60\% Attacked}                                                                                     & \multicolumn{4}{c}{90\% Attacked}                                                                                     \\ \cline{2-16} 
\multicolumn{1}{c|}{}                                                                            & P                         & R                         & F1                         & P                         & R                         & \multicolumn{1}{c|}{F1}    & \multicolumn{1}{c|}{$\Delta atk$} & P                         & R                         & \multicolumn{1}{c|}{F1}    & \multicolumn{1}{c|}{$\Delta atk$} & P                         & R                         & \multicolumn{1}{c|}{F1}    & \multicolumn{1}{c}{$\Delta atk$} \\ \hline
\specialrule{1pt}{0pt}{0pt}
$\text{BERT}_{\textit{noise}}$                                                                                  & 80.64                     & 78.88                     & 79.64                      & 80.67                     & 75.42                     & \multicolumn{1}{c|}{77.17} & -3.10\%                           & 78.44                     & 69.42                     & \multicolumn{1}{c|}{71.33} & -10.43\%                          & 76.46                     & 65.55                     & \multicolumn{1}{c|}{66.96} & -15.92\%                         \\ \hline
$\text{BERT}_{\textit{noise}} + mean$                                                                           & 81.59                     & 77.73                     & 79.18                      & 80.81                     & 73.79                     & \multicolumn{1}{c|}{75.82} & -4.24\%                           & 78.35                     & 68.32                     & \multicolumn{1}{c|}{70.17} & -11.37\%                          & 75.89                     & 65.54                     & \multicolumn{1}{c|}{66.94} & -15.45\%                         \\ \hline
$\text{BERT}_{\textit{noise}} + max$                                                                            & 80.72                     & 79.81                     & \textbf{80.23}                      & 79.68                     & 76.29                     & \multicolumn{1}{c|}{\textbf{77.57}} & -3.31\%                           & 77.46                     & 72.00                     & \multicolumn{1}{c|}{\textbf{73.64}} & -8.21\%                          & 74.82                     & 69.60                     & \multicolumn{1}{c|}{\textbf{71.06}} & -11.42\%                         \\ \hline
$\text{BERT}_{\textit{noise}} + weighted$                                                                       & 81.31                     & 78.06                     & 79.34                      & 80.71                     & 75.10                     & \multicolumn{1}{c|}{76.91} & -3.06\%                           & 77.73                     & 69.73                     & \multicolumn{1}{c|}{71.57} & -9.79\%                          & 75.13                     & 66.77                     & \multicolumn{1}{c|}{68.29} & -13.92\%                         \\ \hline
$\text{BERT}_{\textit{noise}} + first\text{-}last$                                                                          & 81.62                     & 77.63                     & 79.12                      & 80.36                     & 75.04                     & \multicolumn{1}{c|}{76.79} & \textbf{-2.94\%}                           & 78.04                     & 71.48                     & \multicolumn{1}{c|}{73.28} & \textbf{-7.38\%}                          & 74.87                     & 69.15                     & \multicolumn{1}{c|}{70.66} & \textbf{-10.69\%}                         \\ \hline
\specialrule{1pt}{0pt}{0pt}
DeBERTaV3                                                                                  & 82.55                     & 78.70                     & \textbf{80.17}                      & 81.85                     & 74.33                     & \multicolumn{1}{c|}{\textbf{76.48}} & \textbf{-4.60\%}                           & 80.14                     & 68.47                     & \multicolumn{1}{c|}{70.44} & -12.13\%                          & 78.41                     & 63.96                     & \multicolumn{1}{c|}{64.99} & -18.93\%                         \\ \hline
DeBERTaV3$ + mean$                                                                           & 82.19                     & 77.49                     & 79.17                      & 81.22                     & 73.13                     & \multicolumn{1}{c|}{75.28} & -4.91\%                           & 79.83                     & 68.18                     & \multicolumn{1}{c|}{70.10} & -11.45\%                          & 78.31                     & 64.20                     & \multicolumn{1}{c|}{65.31} & \textbf{-17.50\%}                         \\ \hline
DeBERTaV3$ + max$                                                                            & 83.32                     & 76.96                     & 79.02                      & 82.50                     & 72.73                     & \multicolumn{1}{c|}{75.08} & -4.98\%                           & 80.72                     & 67.89                     & \multicolumn{1}{c|}{69.81} & -11.65\%                          & 78.79                     & 63.82                     & \multicolumn{1}{c|}{64.81} & -17.98\%                         \\ \hline
DeBERTaV3$ + weighted$                                                                       & 81.74                     & 77.71                     & 79.21                      & 80.88                     & 73.02                     & \multicolumn{1}{c|}{75.13} & -5.15\%                           & 79.72                     & 68.24                     & \multicolumn{1}{c|}{70.16} & \textbf{-11.42\%}                          & 78.14                     & 64.18                     & \multicolumn{1}{c|}{65.28} & -17.58\%                         \\ \hline
DeBERTaV3$ + first\text{-}last$                                                                          & 82.90                     & 78.32                     & 79.99                      & 81.65                     & 73.75                     & \multicolumn{1}{c|}{75.92} & -5.08\%                           & 79.88                     & 68.71                     & \multicolumn{1}{c|}{\textbf{70.69}} & -11.62\%                          & 78.21                     & 64.54                     & \multicolumn{1}{c|}{\textbf{65.74}} & -17.81\%                         \\ \hline
\specialrule{1pt}{0pt}{0pt}
\end{tabular}
\end{adjustbox}
\caption{Experimental results of offensive language detection when using layer-wise pooling strategies to the $\text{BERT}_{\textit{noise}}$ and DeBERTaV3. The two models differ in whether the texts used for the pre-training stage were preprocessed and their own model structures.}
\label{table_6} 
\end{table*}

%% file: Table_7.tex
\begin{table}[t!]
\centering
\small
\begin{adjustbox}{max width=\textwidth}
\begin{tabular}{l|cl}
\hline
\specialrule{1pt}{0pt}{0pt}
\multicolumn{1}{c|}{\multirow{2}{*}{Model}}                                                      & \multicolumn{2}{c}{Average}                                                \\ \cline{2-3} 
\multicolumn{1}{c|}{}                                                                            & \multicolumn{1}{c|}{F1}             & \multicolumn{1}{c}{$\Delta atk$} \\ \hline
\specialrule{1pt}{0pt}{0pt}
$\text{BERT}_{\textit{clean}}$                                                                                  & \multicolumn{1}{c|}{68.52}          & -12.85\%                         \\ \hline
$\text{BERT}_{\textit{clean}} + max$                                                                            & \multicolumn{1}{c|}{68.51}          & -12.88\%                         \\ \hline
$\text{BERT}_{\textit{clean}} + first\text{-}last$                                                                          & \multicolumn{1}{c|}{\textbf{71.24}} & \textbf{-10.05\%}                \\ \hline
\specialrule{1pt}{0pt}{0pt}
$\text{BERT}_{\textit{noise}}$                                                                                  & \multicolumn{1}{c|}{71.82} & -9.81\%                 \\ \hline
$\text{BERT}_{\textit{noise}} + max$                                                                                  & \multicolumn{1}{c|}{\textbf{74.08}} & -7.64\%                 \\ \hline
$\text{BERT}_{\textit{noise}} + first\text{-}last$                                                                                  & \multicolumn{1}{c|}{73.57} & \textbf{-7.00\%}                 \\ \hline
\specialrule{1pt}{0pt}{0pt}
\begin{tabular}[c]{@{}l@{}}$\text{Ensemble}_{\textit{hard}}$\\ ($\text{BERT}_{\textit{clean}}$ + $\text{BERT}_{\textit{noise}}$)\end{tabular} & \multicolumn{1}{l|}{70.99}          & -11.64\%                         \\ \hline
\begin{tabular}[c]{@{}l@{}}$\text{Ensemble}_{\textit{soft}}$\\ ($\text{BERT}_{\textit{clean}}$ + $\text{BERT}_{\textit{noise}}$)\end{tabular} & \multicolumn{1}{c|}{71.23}          & -11.37\%                         \\ \hline
\specialrule{1pt}{0pt}{0pt}
DeBERTaV3                                                                                  & \multicolumn{1}{c|}{70.63} & -11.88\%                 \\ \hline
DeBERTaV3$ + max$                                                                                  & \multicolumn{1}{c|}{69.89} & -11.53\%                 \\ \hline
DeBERTaV3$+ first\text{-}last$                                                                                  & \multicolumn{1}{c|}{\textbf{70.78}} & \textbf{-11.50\%}                 \\ \hline
\specialrule{1pt}{0pt}{0pt}
\end{tabular}
\end{adjustbox}
\caption{Average from the experimental results of offensive language detection when using max pooling and first-last pooling to the models which exhibited the high scores among the baselines.}
\label{table_7} 
\end{table}

%% file: 06_conclusion.tex
We proposed user-intended adversarial attacks that occur frequently in offensive languages online from the perspective of malicious users. We categorized them into three types: \textsc{Insert}, \textsc{Copy}, and \textsc{Decompose}, which add special symbols or exploit the distinct features of the Korean language. The involvement of attacks significantly affects the tokenization results from the original text.

To address the proposed attacks, we introduced the pooling strategies in a layer-wise manner. This extension utilizes not only the last layer, which focuses on offensiveness, but also the preceding layers, which focus more on token embeddings. The experimental results indicated that first-last pooling was the most robust to the proposed attacks and could even achieve a comparable performance to that of models pre-trained on noisy texts, when applied to models pre-trained on clean texts. We especially demonstrated that rather than the middle layers, the first and last layers can be effectively employed to detect attacked offensive languages.

Furthermore, we experimented with the extent to which the introduced pooling strategies could handle the proposed attacks. We observed that the first-last pooling and max pooling are the most robust, depending on the nature of the texts used for pre-training. It is noteworthy that these strategies, \textit{without} the explicit training of additional parameters or noisy texts, can effectively defend against user-intended adversarial attacks.

%% file: _A-Appendix.tex
\section{User-Intended\\
\;\;\;\;\;\;Adversarial Attacks Details}
\label{appendix_a}

\subsection{\textsc{Insert}}
We utilized three \textsc{Insert} types by adding special symbols that are not complete characters. The detailed methods and examples are as follows:

\begin{itemize}
\item \textsc{Insert}\_\textit{zz}: We randomly added between 2\textasciitilde5 sounds of `ㅋ' to the word (on the keyboard, the sound `ㅋ' corresponds to the alphabet `z'). This is a commonly used expression in online, conveying a meaningless and somewhat frivolous tone. We placed it not only between characters but also in instances where the final sound of a specific character was empty.

\item \textsc{Insert}\_\textit{space}: We randomly added a single space to the word. We expected the same impact as the intentions of malicious users.

\item \textsc{Insert}\_\textit{special}: We randomly added a special character to the word: one of `\textasciitilde', `!', `@', `1', or `2'. We also expected the same impact as the intentions of malicious users.
\end{itemize}

\input{Table_8}

\subsection{\textsc{Copy}}
We utilized three \textsc{Copy} types based on the distinctive characteristics of the Korean language. The detailed methods and examples are as follows:

\begin{itemize}
\item \textsc{Copy}\_\textit{initial}: We copied the initial sound of the character to the final sound of the preceding character. For example, if `기' is chosen from `기레기', the initial sound `ㄱ' would be copied as the final sound to the preceding `레'. Consequently, `레' is transformed into `렉', leading to `기렉기'.

\item \textsc{Copy}\_\textit{middle}: We copied the middle sound of the character, onto the newly added character. If the selected character had a final sound, it was also included. For example, if `딱' is chosen from `틀딱', the middle sound `ㅏ' would be copied as the following character with the final sound `ㄱ'. Consequently, `악' is newly added, leading to `틀따악'.

\item \textsc{Copy}\_\textit{final}: We copied the final sound of the character to the initial sound of the following character. For example, if `있' is chosen from `있었네', the final sound `ㅆ' would be copied as the initial sound to the following `었'. Consequently, `었' is transformed into `썼', leading to `이썼네'.
\end{itemize}

\input{Table_9}

\input{Table_10}

\subsection{\textsc{Decompose}}
We utilized two \textsc{Decompose} types based on the distinctive characteristics of the Korean language. The detailed methods and examples are as follows:

\begin{itemize}
\item \textsc{Decompose}\_\textit{final}: We decomposed the final sound of the character into the newly added sound. For example, if `딱' is chosen from `틀딱', the final sound `ㄱ' would be decomposed as the following sound. Consequently, the expression will be `틀따ㄱ'.

\item \textsc{Decompose}\_\textit{all}: We decomposed all the sounds of the character. For example, if `틀' is chosen from `틀딱', it would be decomposed as the initial, middle, and final sounds. Consequently, the expression will be `ㅌㅡㄹ딱'.
\end{itemize}

\subsection{How the Attack Works}

In our experiments, the attacks were only applied to the test set to evaluate the robustness of the model against user-intended adversarial attacks. We randomly selected one of all the attacks according to the attack rates. For instance, with a 30\% attack rate and 9 words in a sentence, we attacked 3 words (30\% of them), with each word randomly reflecting one of the proposed attacks.

\subsection{Experiments on Each Attack}

We further investigated the impact for each attack type, selecting only one of the attacks from only \textsc{Insert}, \textsc{Copy}, or \textsc{Decompose}. We chose $\text{BERT}_{\textit{clean}}$ and the model that applied first-last pooling, which exhibited the best performance among the layer-wise pooling strategies.

The performances according to the attack types are presented in Table \ref{table_11}. It is noteworthy that a high performance on a specific attack type indicates that it is easier to defend than others, and vice versa. \textsc{Insert} proved to be easier than all attacks, with \textsc{Copy} only marginally harder than \textsc{Insert}. However, \textsc{Decompose} was more difficult than these attacks, exhibiting a performance degradation compared to all attacks. Therefore, we revealed that adversarial attacks that reflect the characteristics of the Korean language are more challenging than those that simply add special symbols that could be adapted language independently.

\input{Table_11}

%% file: Table_8.tex
\begin{table}[h!]
\begin{adjustbox}{max width=\columnwidth}
\centering
\small
\begin{tabular}{l|l}
\begin{tabular}[c]{@{}l@{}}\textsc{Insert} Attacks \end{tabular} & Text Examples                    \\ \hline
\underline{original text}                                                         & \underline{틀딱이냐?} (okay, boomer?) \\
       & \underline{기레기 여기 있었네} (presstitute right here) \\
\textsc{Insert}\_\textit{zz}                                                               & 틀ㅋㅋ딱이냐? \\
       & 기렠ㅋㅋㅋ기 여기 있었네 \\
\textsc{Insert}\_\textit{space}                                                            & 틀 딱이냐? \\
       & 기레 기 여기 있었네 \\
\textsc{Insert}\_\textit{special}                                                          & 틀@딱이냐? \\
       & 기2레기 여기 있었네 \\             
\end{tabular}
\end{adjustbox}
\caption{Text examples of user-intended adversarial attacks with the types of \textsc{Insert}.}
\label{table_8} 
\end{table}

%% file: Table_9.tex
\begin{table}[h!]
\begin{adjustbox}{max width=\columnwidth}
\centering
\small
\begin{tabular}{l|l}
\begin{tabular}[c]{@{}l@{}}\textsc{Copy} Attacks \end{tabular} & Text Examples                    \\ \hline
\underline{original text}                                                         & \underline{틀딱이냐?} (okay, boomer?) \\
       & \underline{기레기 여기 있었네} (presstitute right here) \\
\textsc{Copy}\_\textit{initial}                                                            & 틀딱인냐? \\
       & 기렉기 여기 있었네 \\
\textsc{Copy}\_\textit{middle}                                                             & 틀따악이냐? \\
       & 기레기 여기이 있었네 \\
\textsc{Copy}\_\textit{final}                                                              & 틀딱기냐? \\
       & 기레기 여기 이썼네 \\             
\end{tabular}
\end{adjustbox}
\caption{Text examples of user-intended adversarial attacks with the types of \textsc{Copy}.}
\label{table_9} 
\end{table}

%% file: Table_10.tex
\begin{table}[b!]
\begin{adjustbox}{max width=\columnwidth}
\centering
\small
\begin{tabular}{l|l}
\begin{tabular}[c]{@{}l@{}}\textsc{Decompose} Attacks \end{tabular} & Text Examples                    \\ \hline
\underline{original text}                                                         & \underline{틀딱이냐?} (okay, boomer?) \\
       & \underline{기레기 여기 있었네} (presstitute right here) \\
\textsc{Decompose}\_\textit{final}                                                         & 틀따ㄱ이냐? \\
       & 기레기 여기 이ㅆ었네 \\
\textsc{Decompose}\_\textit{all}                                                           & ㅌㅡㄹ딱이냐? \\
       & 기ㄹㅔ기 여기 있었네 \\             
\end{tabular}
\end{adjustbox}
\caption{Text examples of user-intended adversarial attacks with the types of \textsc{Decompose}.}
\label{table_10} 
\end{table}

%% file: Table_11.tex
\begin{table}[h!]
\begin{adjustbox}{max width=\columnwidth}
\begin{tabular}{l|ccc|ccc}
\hline
\specialrule{1pt}{0pt}{0pt}
\multicolumn{1}{c|}{\multirow{2}{*}{Model}} & \multicolumn{3}{c|}{All Attacks}                         & \multicolumn{3}{c}{Only \textsc{Insert}}    \\ \cline{2-7} 
\multicolumn{1}{c|}{}                        & P                 & R                 & F1               & P                  & R                  & F1                 \\ \hline
\specialrule{1pt}{0pt}{0pt}
$\text{BERT}_{\textit{clean}}$                        & 77.74             & 66.38             & 67.96            & 78.12              & 68.80              & 70.66              \\
$\text{BERT}_{\textit{clean}} + first\text{-}last$                & 77.58             & 69.38             & 71.21            & 77.51              & 70.76              & 72.54              \\ \hline
\specialrule{1pt}{0pt}{0pt}
\multicolumn{1}{c|}{\multirow{2}{*}{Model}} & \multicolumn{3}{c|}{Only \textsc{Copy}}                  & \multicolumn{3}{c}{Only \textsc{Decompose}} \\ \cline{2-7} 
\multicolumn{1}{c|}{}                        & P                 & R                 & F1               & P                  & R                  & F1                 \\ \hline
\specialrule{1pt}{0pt}{0pt}
$\text{BERT}_{\textit{clean}}$                        & 77.86             & 68.47             & 70.29            & 78.38              & 62.23              & 65.67              \\
$\text{BERT}_{\textit{clean}} + first\text{-}last$                & 77.33             & 70.50             & 72.26            & 77.00              & 66.47              & 68.04              \\ \hline
\specialrule{1pt}{0pt}{0pt}
\end{tabular}

\end{adjustbox}
\caption{Experimental results of offensive language detection when the attack ratio is 60\% for each type of user-intended adversarial attacks.}
\label{table_11}
\end{table}

%% file: _B-Appendix.tex
\section{Experimental Details}
\label{appendix_b} 

\subsection{Implementation Details}
We used 6 layers with an embedding dimension of 768 and a dropout ratio of 0.1 for the RNN-based models containing BiLSTM and BiGRU. As for the BERT-based models, containing $\text{BERT}_{\textit{clean}}$, $\text{BERT}_{\textit{noise}}$, their ensembles, and DeBERTaV3, we fine-tuned 12 pre-trained layers with an embedding dimension of 768 and a dropout ratio of 0.2.

We used the AdamW optimizer with a learning rate of 1e-5, trained the models for 1\textasciitilde5 epochs, and considered the epoch with the lowest validation loss or the last epoch. We set a batch size of 32 for all the models. The models were implemented using PyTorch and NVIDIA GeForce RTX 3090 GPU. We also used the HuggingFace library to leverage the weights of the pre-trained BERT models.

\subsection{Metrics}
We collected the datasets to train as many types of offensive languages as possible. There is some label imbalance, therefore, we used macro precision, recall, and f1-score to address this issue.

$\Delta atk$ represents the performance degradation in the f1-score. For example, denoting the existing f1-score as $\text{F1}_{\textit{original}}$ and the f1-score with the attacks as $\text{F1}_{\textit{attacked}}$, it is computed as follows:
\begin{align}\label{eq5}
\Delta atk &= (\text{F1}_{\textit{attacked}} - \text{F1}_{\textit{original}}) / \text{F1}_{\textit{original}} * 100.
\end{align}

%% file: _C-Appendix.tex
\section{Examples of the Attacks}
\label{appendix_c}

We presented the following cases where offensive languages contained user-intended adversarial attacks on popular online communities in South Korea, such as Twitter\footnote{https://twitter.com} and dcinside\footnote{https://dcinside.com}.

\small
\begin{itemize}
    \item \textsc{Insert}\_\textit{zz}: \\ 지에지에 개 병ㅋㅋㅋ신 \\ (ziezie so dumbzzzfuck)
    
    \item \textsc{Insert}\_\textit{space}: \\ 완전 또 라이 아니야... \\ (crazy ass we irdo)
    
    \item \textsc{Insert}\_\textit{special}: \\ 넥슨아 이거 개같으면 지워라 새@끼들아 \\ (delete this if you dare Nexon you sons of bit@ches)

    \item \textsc{Copy}\_\textit{initial}: \\ 우리나라 남자샊끼들 진짜 재미없게 산다 \\ (korean lossser men live in the most boring ass lives)

    \item \textsc{Copy}\_\textit{middle}: \\ 역쉬나 기레에기들.. \\ (presstiiitutes of course)

    \item \textsc{Copy}\_\textit{final}: \\ 현생 어떻게 살아가누 너가튼게? \\ (how the fuck dooyou live in the real world?)

    \item \textsc{Decompose}\_\textit{final}: \\ 과몰입 중인 병신 트ㄹ딱들ㅋㅋ \\ (chronically online dumbass bo omers lmao)

    \item \textsc{Decompose}\_\textit{all}: \\ 너 같은 ㄱㅐ 새끼는 몽둥이로 맞아 죽어야 해 \\ (p i ece of shit like you need to be beaten to death)
\end{itemize}